# Towards a framework for evaluating the safety, acceptability and efficacy of AI systems for health: an initial synthesis

Jessica Morley[1,2], Caroline Morton[2], Kassandra Karpathakis[3], Mariarosaria Taddeo[1,4], Luciano Floridi[1,4]

1. Oxford Internet Institute, University of Oxford, 1 St Giles', Oxford OX1 3JS
2. Nuffield Department of Primary Care Health Sciences, University of Oxford, Radcliffe Primary Care Building, Radcliffe Observatory Quarter, Woodstock Rd, Oxford OX2 6GG
3. NHSX, Skipton House, 80 London Road, SE1 6LH
4. Alan Turing Institute, The British Library, 2QR, 96 Euston Rd, London NW1 2DB

Correspondence author: Jessica.morley@oii.ox.ac.uk

## Abstract

The potential presented by Artificial Intelligence (AI) for healthcare has long been recognised by the technical community. More recently, this potential has been recognised by policymakers, resulting in considerable public and private investment in the development of AI for healthcare across the globe. Despite this, excepting limited success stories, real-world implementation of AI systems into front-line healthcare has been limited. There are numerous reasons for this, but a main contributory factor is the lack of internationally accepted, or formalised, regulatory standards to assess AI safety and impact and effectiveness. This is a well-recognised problem with numerous ongoing research and policy projects to overcome it. Our intention here is to contribute to this problem-solving effort by seeking to set out a minimally viable framework for evaluating the safety, acceptability and efficacy of AI systems for healthcare. We do this by conducting a systematic search across Scopus, PubMed and Google Scholar to identify all the relevant literature published between January 1970 and November 2020 related to the evaluation of: output performance; efficacy; and real-world use of AI systems, and synthesising the key themes according to the stages of evaluation: pre-clinical (theoretical phase); exploratory phase; definitive phase; and post-market surveillance phase (monitoring). The result is a framework to guide AI system developers, policymakers, and regulators through a sufficient evaluation of an AI system designed for use in healthcare.

**Funding**

Jessica Morley was partially supported by a European AI Policy Network Fellowship 2020-21, which was funded by Google EU. JM is also supported by a Wellcome Trust Doctoral Fellowship.

## 1. Introduction

The potential presented by Artificial Intelligence (AI) for healthcare was first recognised by the technical community in 1959 (Ledley & Lusted, 1959). Indeed, by 1995 there were approximately 1000 citations of neural networks (one branch of AI) in the biomedical literature (Lisboa, 2002). More recently, in the past five-to-ten years, this potential has been recognised by policymakers, resulting in considerable public and private investment in the development of AI for health across the globe. Despite this, excepting limited success stories, real-world implementation of AI systems into front-line healthcare has been limited. There are numerous reasons for this slow rate of adoption, including the fact that there are considerable barriers to using healthcare data for even basic routine analysis (Bacon & Goldacre, 2020). However, the primary reasons are likely to be ethical, social and legal (Greenhalgh et al., 2017).

Healthcare practitioners must abide by the principle: *do no harm* (Pellegrino, 1990). According to the practice of evidence-based medicine, this at least partially means ensuring that the tools they use for diagnosis, prognosis and treatment are provably clinically *efficacious* and *safe* (Sackett & Rosenberg, 1995). Typically, this involves reviewing evidence of efficacy derived from (ideally) randomised controlled trials (RCTs), and regularly reviewing data from post-market surveillance. Admittedly, the process is not without problems, and it has been rightfully criticised for being too positivistic and oblivious of the importance of other aspects of care, such as the values and needs of the clinician or the patient (Greenhalgh et al., 2014). However, it remains true that this standardised generation of evidence mitigates risks to patient safety and therefore increases clinician confidence in the intervention. The challenge presented by AI systems is that, whilst RCTs have come to be widely recognised as the most reliable method of determining effectiveness, they have traditionally been used to evaluate the effects of a single intervention, such as a specific drug (Campbell, 2000). AI systems are inherently more complex. They can be difficult to explain and can be expected to auto-update after an initial review or trial period (Reddy et al., 2020); they usually involve elements of human-computer interaction which can make reproducibility harder (Lisboa, 2002); and the clinical problems addressed by AI solutions often lack a clear gold standard, limiting the applicability of a traditional RCT (Ben-Israel et al., 2020). In short, the complexity of AI solutions presents many considerable challenges for the standard process of defining, developing, documenting, and reproducing involved in evaluating the effectiveness of a clinical intervention (Campbell, 2000). These challenges mean that regulatory standards to assess AI safety, impact and effectiveness have yet to be formalised in most countries (Reddy et al., 2020)[1] and consequently, very few AI solutions have been externally validated in a way that can predict real-world impact (Hopkins et al., 2020). Concerningly, this lack of evidence generation is presenting a barrier to adoption and commissioning, and enabling unsafe deployment practices (Reddy et al., 2020). Fortunately, this does not have to be the case.

There are currently no *formalised* or *standardised* means of evaluating the effectiveness of AI in healthcare. However, in the technical literature – particularly in the context of expert systems and rule-based clinical decision support – the evaluation of AI systems has always been an important part of their implementation (Miller, 1986). Shortliffe & Davis, for example, set out an eight-stage process for the implementation of knowledge-based clinical decision support software[2] in (1975):

1. Define system's long-range goals;

2. Develop system with frequent re-evaluations to assess design adequacy;

3. Formally evaluate system performance;

4. Formally evaluate system acceptability and efficacy in a prototype 'real world' environment;

5. Review service functionality over an extended period in the prototype environment;

6. Conduct follow-up studies to demonstrate the system's large-scale usefulness;

7. Make final changes needed to allow for wide system distribution;

8. Release system onto the market alongside appropriate plans for maintenance and updating.

Indeed, by the 1980s the need for rigorous evaluation of the quality and impact of AI was well recognised. Numerous papers were published at the time considering how to evaluate system performance in laboratory and real-world settings, factoring in complexities associated with legal and ethical issues; system transferability; brittleness; and the dynamic nature of healthcare knowledge. By the end of the period, it was established (Magrabi et al., 2019) that the evaluation of a health-related AI system must involve:

- testing the validity of the system i.e., its correctness in reasoning;

- the usefulness for clinical work; and

- the effects and impacts of the system on clinical work, care processes, and patient outcomes.

---

[1] This is based on an in-depth policy analysis conducted of documents in the public domain identified using the same search terms outlined in the methodology section. It is, of course, entirely possible that there are policy in development that are not yet in the public domain.

[2] Clinical Decision Support Software (typically referred to as CDSS) is defined by the US Food and Drug Administration as "software that is intended to provide decision support for the diagnosis, treatment, prevention, cure, or mitigation of diseases or other conditions" (FDA, 2019)

Of course, in the intervening years, AI techniques have developed and moved beyond rule-based decision-making systems, trained expert-systems, and models largely based on regression, to more complex techniques related to unsupervised and reinforcement learning. The exponential growth in computational power and the size of available databases has also transformed the nature of several sub-sectors of AI, from machine learning to natural language processing. Additionally, there has been a realisation that AI systems are not 'just' statistical models but sociotechnical systems with considerable agency that can cause significant harm. Therefore their evaluation must consider the entire human, social and organisational infrastructure from which they emerge and are embedded in (Macrae, 2019). These additional considerations are reflected, at least partly, in more recent work, including the CONSORT-AI extension – which provides guidelines for clinical trial reports involving AI (Liu et al., 2020) – and proposals developed by the Japanese and American regulatory bodies responsible for medical devices (B. Allen, 2019; Ota et al., 2020). However, what is still lacking is a critical synthesis of these different bodies of literature. This is the task we set ourselves in the following pages, as we seek to create a minimally viable framework for evaluating the safety, acceptability and efficacy of AI systems for healthcare. Specifically, section two provides the methodology; section three contains a brief summary of the systematic search results; section four provides a discussion of the framework; section 5 outlines a series of policy recommendations for further developing and testing the framework; and section 6 concludes the article.

## 2. Methodology

Typically, in the field of evidence-based medicine and evidence-based policy, synthesising research is an *integrative* process, involving the summarisation of (primarily) quantitative data from primary studies, for example, in a meta-analysis (Dixon-Woods et al., 2005). Research of this nature is valid but is most useful when the concepts or variables to be included in the synthesis are well specified and/or when the intended output needs to have an element of causality. The utility of these so-called "rationalist-approaches" is reduced when dealing with sociotechnical systems because the dynamic elements of such systems are poorly captured by quantitative data. Thus, in this instance, we opted to take what is known as a "rhetorical-interpretivist" approach to synthesis (Greenhalgh & Russell, 2006).

Interpretive syntheses can include all types of data, both qualitative and quantitative, and are typically conceptual in process and output, rather than quantitative (Schloemer & Schröder-Bäck, 2018). Thus, like Schloemer and Schröder-Bäck (2018) in their development of 'criteria for evaluating transferability of health interventions', we conducted a thematic synthesis (a form of interpretive synthesis) to identify the key concepts involved in the evaluation of AI systems and combined them in a higher-order theoretical structure through a process of induction and interpretation.

First, we conducted a systematic search across Scopus, PubMed and Google Scholar using the search terms in Box 1 to identify all the relevant literature published between January 1970 and November 2020 related to the evaluation of: output performance; efficacy; and real-world use of AI systems. The results of this search are discussed in the next section. Papers were first screened by title, then by abstract and then by full text against the eligibility criteria set out in Box 2, with some assistance from members of the NHSX AI Lab. For each of the papers reviewed in full, we extracted the information listed in Box 3.

AI|ML|algorithm|"expert system" & efficacy & healthcare

AI|ML|algorithm|"expert system" & evaluation & healthcare

AI|ML|algorithm|"expert system" & verification & healthcare

AI|ML|algorithm|"expert system" & verification & healthcare

AI|ML|algorithm|"expert system" & implementation & healthcare

AI|ML|algorithm|"expert system" & regulation & healthcare

AI|ML|algorithm|"expert system" & Governance & healthcare

AI|ML|algorithm|"expert system" & accuracy & healthcare

AI|ML|algorithm|"expert system" & sensitivity & healthcare

AI|ML|algorithm|"expert system" & safety & healthcare

Box 1: Terms used to search Scopus, PubMed and Google Scholar

**Inclusion Criteria:**

Paper provides specific description of method to evaluate AI system intended for use in frontline care

Paper provides specific description of AI system evaluation criteria for use in frontline care

Paper provides a case study of AI system evaluation for use in frontline care

Paper provides a specific description of AI system safety criteria for use in frontline care

**Exclusion Criteria:**

Paper provides purely theoretical/narrative description of the need to evaluate AI systems

Paper provides a description of evaluation method/criteria that is not generalisable

Paper provides a description of an evaluation method for AI system sold directly to consumer e.g., in an 'app'

Box 2: Inclusion/ Exclusion criteria used for reviewing papers returned by search

Type of article

Publication year

Location of first author affiliation

Evaluation criteria listed/used in evaluation, for example, positive predictive value, bias, accuracy, sensitivity, specificity, pooled sensitivity etc.

Safety criteria if these are mentioned, for example, reporting of adverse incidents

Methods of evaluation mentioned for example, test-train, RCT, observational trial, compare human to machine

| Notes on how to implement the evaluation in practice |
| --- |
| Factors to consider, for example, does the article split the results by type of algorithm, stage of development etc. |

Box 3: Information extracted for each of the included articles

Second, we used iterative coding analysis to categorise the themes covered in each of the fully-reviewed papers according to the stages of evaluation set out by Lisboa (2002): pre-clinical (theoretical phase); exploratory phase; definitive phase; and added the post-market surveillance phase (monitoring) typically needed for long-term safety monitoring. To inform our interpretation, we relied on our own experiences with software development, evaluation, and previous discussions with developers of AI systems. This interpretivist approach was similarly used by Sendak et al., (2020) to great effect in their narrative review proposing a general framework for translating machine learning into healthcare.

Finally, we added to the thematic analysis relevant insight from complexity science, implementation science and social science (Greenhalgh & Papoutsi, 2019) to ensure that appropriate consideration was also given to institutions, interactions between agents, and social processes (Alford, 1998). The result is a framework that can guide AI system developers, policymakers, and regulators through a sufficient evaluation of an AI system designed for use in healthcare. The aim is to offer a framework that can give to those responsible for implementing AI systems into healthcare settings a satisfactory level of confidence that the system is accurate, precise, safe and usable. A discussion detailing the contents of this framework follows a brief analysis of the search results.

## 3. Search Results

Figure 1 summarises the results of the search process. The original searches returned a total of 6,228 results, of which 41 papers were included in the final review. Details of the included papers are summarised in Appendix 1. The papers covered the time period 1986 – 2020; they came from the European Union (EU), United Kingdom (UK), United States of America (USA), Canada, Asia-Pacific and the United Arab Emirates; and they were primarily other review papers. A more detailed breakdown is provided in Table 2. Review papers written during the 2010s from US institutions were the most common. As will become clear in the discussion section, the greatest number of papers focused on 'validation' or the exploratory stage of evaluation.

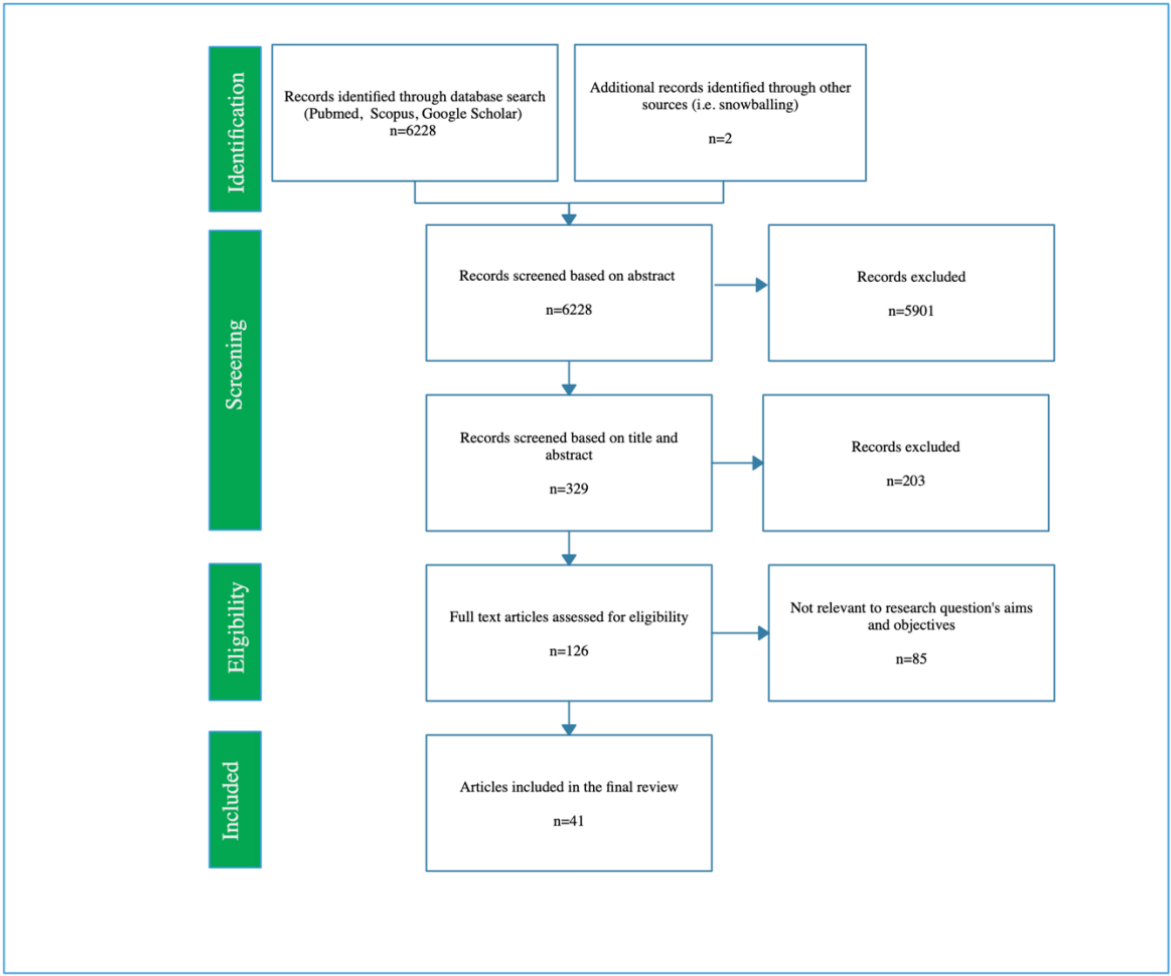

Figure 1: PRISMA flow-chart of search results

| Descriptive Characteristic | Number of Included Papers |
|---|---|
| *Decade of publication* | |
| 1980s | 1 |
| 1990s | 2 |
| 2000s | 7 |
| 2010s | 17 |
| 2020s | 14 |
| *Country of Lead Author Affiliation* | |
| EU | 8 |
| UK | 9 |
| USA | 14 |
| Asia-Pacific | 6 |
| UAE | 1 |
| International | 1 |
| *Type of Paper* | |
| Case Study | 4 |
| Discussion paper | 6 |
| Protocol | 2 |
| Review | 29 |

Table 1: Summary characteristics of included papers

## 4. Analysis

What follows is a brief overview of the results for each of the four phases outlined in the methodology. The full framework is provided at the end of this section.

### 4.1 Preclinical Phase (Verification Stage)

The preclinical phase is primarily intended to set out a clear rationale for why an AI system is suitable for the clinical problem at hand (Lisboa, 2002). It is essentially designed to answer the question: 'is this the right solution to the problem?', or, to borrow a technical term from software engineering, to complete the *verification* process and ensure that the planned AI system is capable of satisfying all the expected requirements. More specifically those conducting this evaluation phase should, upon completion, feel satisfied that the proposed AI system is both technically feasible and likely to be socially acceptable.

To ensure that the proposed system meets these criteria, those producing evidence for evaluation (most likely the AI developers) should combine quantitative methods, such as modelling, with qualitative methods (for example, focus groups, preliminary surveys or case study analysis) to identify the relevant system components, understand how they interact, and ultimately produce clearly defined clinical outcomes (Campbell, 2000). These outcomes should be defined, preferably in public, before system development begins, so that those responsible for developing the system can be held accountable for ensuring their system meets these and no other (potentially) more convenient outcomes. Measures such as this, can ensure that the whole process of AI system development abides by the principle of meticulous transparency (Benrimoh et al., 2018).

Transparency about the development process can help to ensure trust and hence acceptance from patients, the public, clinicians, and where appropriate - commissioners. However, transparency is not, in itself, sufficient to ensure trust if these key stakeholders feel as though there is a disconnect between the technical 'AI system' and the wider sociotechnical health system within which it is embedded. Evaluations which ignore the complexities generated by the interactions between individual agents, their decisions and actions, their motivations, values and professional norms (May, 2013) typically fail (Greenhalgh & Russell, 2010). Thus, complexities should be identified from the start so that mitigating actions can be taken (Greenhalgh et al., 2017). This will require AI system developers to engage in activities, such as service or care pathway mapping to gain a deeper understanding of the wider service context, and the boundaries of the AI system's influence (Magrabi et al., 2019). The production of this will help evaluators detect mechanisms that might lead to deviations from expectations when the system is implemented and help developers plan for this level of dynamism (Greenhalgh et al., 2018).

## 4.2    *Exploratory Phase (Validation Stage)*

The purpose of the exploratory phase is to develop and statistically assess the AI system to ascertain its performance compared to existing benchmarks, and ensure that its results are reproducible in different settings (Lisboa, 2002). To borrow another technical term from software engineering, this is the *validation* stage where the key question for evaluators is whether the AI system has been designed in the right way.

Miller set out the basic key components for this phase in 1986, stating that validation of an AI system needs to cover: accuracy; completeness and consistency; performance compared to baseline (clinician or previously validated system); existence of bias; and causes of failure. Those reviewing evidence for evaluation at this stage should be assured that the AI system is functional; that it at least matches human or expert-level performance; and that the system can cope with data that accurately and fully reflect the expected variability in the intended usage setting (England & Cheng, 2019). The general consensus is that this evidence should be generated statistically, with AI developers using appropriate statistical tests to evaluate predictive accuracy; sensitivity; specificity;

generalisability; reliability; precision; and repeatability (Attardo et al., 2020; Awaysheh et al., 2019; Choudhury & Asan, 2020; Emmert-Streib et al., 2019; Gottliebsen & Petersson, 2020; Hagiwara et al., 2020; Park & Han, 2018; Reisman, 1996; Shen et al., 2019). The exact method of statistical validation used by evaluators will depend on the type of model used by the AI system. For example, Handelman et al. (2019) state that classification models (e.g., those that predict the presence or absence of a disease) are typically evaluated with the receiver operating characteristic curve (ROC curve) or confusion matrix, whereas regression models (e.g., those predicting outcomes using continuous variables) are more typically evaluated with mean squared error, mean absolute error, or the coefficient of determination (r-squared). The exact 'optimum' performance for each of these statistical tests will also vary depending on the model, the clinical setting, and crucially the clinical outcomes it aims to achieve (Gilvary et al., 2019). For example, as Magrabi et al., (2019) discuss, an AI system that is used for triage will need high discrimination, whereas one used to predict risk in shared decision-making will need to be highly accurate and precise for all types of patients. This is why it will be necessary to define with clinical end-users the precise definition of 'correctness' and optimum performance ahead of time (Miller, 1986).

Once decisions have been made on the appropriate methods of statistical evaluation, the validation process should be built up in three phases. First, the AI system should be validated against the 'test' data of the training dataset in a process of internal cross-validation (Lisboa, 2002). Second, the AI system should be validated against *at least* one external dataset including data from a different patient population than the one the system was trained on – either because it comes from a different clinical setting or the same clinical setting but a different time period (Hopkins et al., 2020). Third, the AI system should be subject to prospective observational validation by exposing it to data representative of the diversity and variation in real-life clinical practice (i.e. datasets that come from multiple clinical centres) (Hopkins et al., 2020). Guidelines for this process are being produced by the upcoming TRIPOD-AI (Collins & Moons, 2019) statement, the STARD-AI statement (Sounderajah et al., 2020) and DECIDE-AI (The DECIDE-AI Steering Group, 2021). Ideally, this third phase would be followed by a prospective randomised controlled trial, but there is general acceptance in the literature that might not always be possible. Within each phase, performance of the AI system should be compared to the gold standard performance of either human or previously validated expert system on the relevant dataset. Based on the outcomes of each sub-evaluation, the AI system developers should go back to the feature engineering to manipulate the features and increase their performance according to the predetermined definition of optimum performance (Angehrn et al., 2020). Although seemingly complex, this process will ensure that the claims of the AI system developers are validated using novel datasets reflecting demographic diversity before marketing the product for clinical practice (Allen et al., 2019).

It is important to note that, at the end of this technical validation process, the AI system might be mathematically optimal but still ethically problematic (Magrabi et al., 2019) for a number of reasons that have been discussed at length elsewhere (Morley, Machado, et al., 2020). This is why technical validation is not sufficient. AI systems should also be subject to ethical validation (more on this in the next section). Ideally, this would take the form of ethics-based auditing (Mökander et al., n.d.), but, at least, a sufficient explanation for the AI system should be produced to check for issues with bias, data quality, and more. Not only will this provide a minimum viable ethical check, but it will also ensure that the AI system is compliant with the legal doctrine of learned intermediaries, which states that clinicians must be able to understand the operation of the system well enough to be able to take responsibility for the results of its use (Lisboa, 2002).

### 4.3    *Exploratory Phase (Real-World Evaluation Stage)*

This third phase, the exploratory or real-world evaluation stage, is more subjective than the previous phases. The aim is to test the clinical efficacy of the system by means of routine use (Campbell, 2000), within a limited number of sites, *before* its deployment is scaled up (Miller, 1986). Evaluators should focus on comparing outcomes before and after the AI system's implementation and seek to assure themselves of the system's clinical efficacy, utility and safety by addressing questions such as these (set out by (Magrabi et al., 2019)): is the system operationally meaningful and useful? Does it fit clinical workflow? Does it still represent up-to-date clinical knowledge? Does its use change clinical decisions? How confidence do clinicians feel in using it? Can automation bias be managed? In short, the key underlying question is whether the AI system is working in the right way.

To answer this underlying question, the focus throughout this stage of evaluation should be on how well the AI system integrates with existing clinical practice (Angehrn et al., 2020); how effectively it achieves its intended purpose (Milne-Ives et al., 2020); and how performance in these areas can be affected by end-user satisfaction (clinician, or even patient, satisfaction) with aspects of software design such as user friendliness, documentation, reliability, and response time (Rivers & Rivers, 2000). As before, the exact performance metrics should be decided with the users of this specific system, rather than these being imposed top down. This is because, ideally, these performance metrics would align with the values, priorities and vision of the organisation(s) in which the AI system is deployed. Misalignment with values can result in non-adoption or abandonment of new technologies (Greenhalgh et al., 2017). Chen & Decary (2020) do provide a general guide, arguing that performance metrics should consider whether clinical effectiveness has been improved in terms of quality, efficiency, and safety; whether access has been expanded for patients; whether patient experiences or operational processes have been improved; whether staff satisfaction has increased; and whether the AI system delivers value for money. A similar list is provided by Simantirakis et al., (2009).

Unlike the preceding statistical phase, there is limited consensus in the literature about *how* this more qualitative evaluation should be conducted. Ben-Israel et al., (2020) suggest that, until an appropriate study design framework is developed, evaluators may need to rely on cohort studies where outcomes prior to, and following, the implementation of an AI system are compared. They note that, whilst this approach will have issues with bias, it will at least provide an estimate of the performance of the system in clinical practice. Case study analysis might also work well, combining documentary analysis with in-depth interviews to understand the social and cultural limitations to the AI system as well as the technical limitations.

The final output of this phase should be the publication of a detailed safety case, which explains with evidence how risks to patient safety have been managed from an organisational perspective. There should be a process built around this safety case, which states when it will be updated, and how adverse event reporting will be handled (Macrae, 2019). This should align with relevant national legislation around medical device reporting.

## 4.4    *Ongoing Monitoring*

This final phase is not bounded in the same way as the previous phases. Instead it is an ongoing post-market surveillance process, much like that which already exists for traditional pharmaceutical interventions (Hopkins et al., 2020). The inclusion of this phase in the framework is a consequence of the fact that the implementation of an AI system does not have a definitive endpoint. As Sendak et al., (2020) note, data quality, population characteristics, and clinical practice all change over time in ways that can impact the validity and utility of AI systems. In addition, some AI models – particularly those using machine learning – are dynamic in the sense that they have the capacity to react to and learn from new data in a way that changes model performance. This can cause safety issues if left unmonitored. Regular monitoring is, therefore, required to ensure that the model remains reliable and safe for use in clinical care (Chen & Decary, 2020).

This regular monitoring should include re-evaluation of the performance metrics outlined in the exploratory/validation phase at pre-specified intervals, and also means of monitoring for issues related to safety, detailed extensively by Challen et al., (2019) and Ellahham et al., (2020). Such issues include: distributional shift; automation bias; reward hacking; and reinforcement of outmoded practice. In addition, regular qualitative assessment should also be conducted to capture changes in user satisfaction (Magrabi et al., 2019), or trust, and to capture variations in patient reported outcomes. Data collected during this process can be used to inform corrective plans, such as recalibrating the model, or training for clinicians (Hopkins et al., 2020). If it is not possible to develop a mechanism for monitoring the performance of dynamic systems, then it may be that 'sunset' clauses are needed, whereby systems are licensed only for a short period of time before needing to be revalidated by going back to phase 2.

Finally, there should be a regular comparison between the actual and stated outcomes of the AI system model. There should be a clear definition of 'failure' set out, which provides acceptable parameters for deviation from optimum performance. Crucially, a plan should be put in place for safely removing the AI system from the relevant service or care if it starts failing, causing patient safety issues, or consistently underperforming (Morley, Cowls, et al., 2020). This plan should also be enacted if, at any point, technical support for the AI system is removed, for example if the providing company goes out of business. This is reflected in the full framework set out below in Box 4.

---

**Framework for evaluating the safety, acceptability and efficacy of AI systems for health**

**Preclinical Stage (Theoretical Stage)**

*To borrow from software engineering this is the 'verification' stage where the key question is: is this AI system the right solution to the problem?*

- Has the clinical problem that the AI system is seeking to target through diagnosis, triage, prognosis, prediction or treatment been defined?
- Has the purpose of the AI system, the health outcomes it aims to achieve, and the metrics that will be used to measure performance, been clearly articulated?
- Has a service/care pathway mapping exercise been conducted to identifying the elements of the service that the AI system will interact with, and to identify the boundaries of the service?
- Have key stakeholders and end users been identified for in-depth user research?
- Has a plan for conducting user research been outlined?
- Have the information needs of end-users who will be interacting with the system been outlined and tested with end-users?
- Has the appropriate 'gold standard' benchmark (or current standard of care where no gold standard exists) against which the AI system's performance will be compared been identified?
- Have the AI system developers checked for already existing automated solutions and reviewed performance of these, including the performance of simpler models e.g., linear regression?
- Have the AI system developers assessed sources of complexity and identified possible means of reducing it?

- Where possible has the implementation of the system and the impact on the service/care pathway been modelled?

**Exploratory Stage (Validation Stage)**

*To borrow from software engineering this is the 'validation' stage where the key question is: has the AI system been designed in the right way?*

- Is there sufficient evidence that training data is of sufficiently high quality?
- Has a clear rationale for choosing which performance metrics e.g., discrimination vs. sensitivity and specificity to optimise been set out?
- Have the appropriate statistical validation methods for testing the following performance metrics been identified? :
  - Accuracy
  - Sensitivity
  - Specificity
  - Generalisability
  - Reliability
  - Precision
  - Repeatability
  - Robustness
- Has a three-phase validation process been conducted? Are the results available for public review for in silico validation; out-of-sample validation; prospective observational validation?
- Is there sufficient evidence that at each validation phase, performance of the model was compared to that of the gold standard on the same datasets?
- Has a sufficient explanation of the model been produced?
- Has the model been subject to an ethics-based audit?
- Has a timeline for regular re-validation been set out?

**Definitive Stage (Real-world Evaluation Stage)**

*This is the 'evaluation' stage and involves real-world testing, the primary question is: is the AI system working in the right way?*

- Has a plan been put in place to evaluate the following?
  - Impacts on key performance metrics when AI system is dealing with missing or messy data
  - Impact on key performance metrics before/after system update
  - Difference between expected and actual performance of system
- Has the impact of the AI system on key outcomes been appropriately assessed? Do the reported outcomes match those pre-specified?

- Has a cohort study or case study of impact been completed?

- Has an audit of patient safety/ adverse events before and after system implementation been completed?

- Has end-used satisfaction been assessed?

- Has it been established that the AI system is sufficiently usable and interpretable?

- Has the impact of the AI system's implementation on the overall service/ care pathway been assessed? For example, has the impact on workflow been considered?

- Is there sufficient evidence that end-users trust the system and that there is a plan in place for dealing with automation bias?

- Has a safety case been produced and published?

**Post-Market Surveillance**

***To borrow from the development of new medical devices this is the stage focused on safety and adverse event monitoring, the main question is: is the AI system having the right kind of impact***

- Has a timeline been set out for revalidation? And re-evaluation?

- Is there a plan in place for the removal of the AI system in case of failure?

- Is there an appropriate mechanism in place for reporting patient safety issues?

- Have appropriate measures been established to monitor for safety issues related to technical problems, such as distributional drift and reward hacking?

Box 4: Framework for evaluating the safety, acceptability and efficacy of AI systems for health

## 5. Discussion and Recommendations

The previous synthesis has produced what can be seen as a minimally viable framework for evaluating the safety, acceptability and efficacy of AI systems designed for use in frontline clinical care settings. It has shown that, whilst no formally standardised framework of this kind exists in official international policy (at least not yet), this is not for lack of knowing *what should* be done. Like with other aspects of pro-ethical AI design (Floridi, 2016), and indeed with healthcare management, the design challenge presents itself in the operationalisation stage or in the moving from the *what* to the *how* (Morley, Floridi, et al., 2020). This is a complex challenge, but not an insurmountable one. There are concrete steps that policymakers, at a national and supranational level, can take to start to turn this high-level framework into actionable policy.

First, there is often a gap between what is *preferable* and what is *feasible* when it comes to technology policy design. The framework very much represents what is *preferable*. However, it is also comprehensive, and it may not be *feasible,* or indeed *reasonable,* to ask AI developers to subject their products to such a rigorous evaluation at least not without additional resource provided by central government, for example. Thus, the practicality of the framework must be rigorously tested before it can be implemented. It should be workshopped with all those involved in developing AI systems including data scientists, product managers, and software engineers. The framework should be iterated in response to feedback gathered through appropriate mechanisms, including consultation and focus groups, and the final set of criteria included should be ratified through a formalising process, such as a Delphi study.

Second, to reduce the risk of duplication and to assess how such an evaluation framework could be embedded in practice, it will be necessary to identify which existing bodies are responsible for conducting which stages of the evaluation and which parts of the evaluation might already be covered by existing Governing processes. A mapping exercise should therefore be conducted to align stages and tasks of the framework with stages and tasks of, for example, medical device or ISO standard certification. A gap analysis should then be conducted to see which aspects of the framework are not yet covered by existing processes. The work by Larson et al. (2020) on regulating image-recognition algorithms provides an excellent example of how to do this.

Third, consideration should be given to situations in which the application of the framework may vary. For example, differences in the level of risk associated with the clinical task that is being automated by the AI system; maturity of the system design; clinical setting; and clinical domain (e.g., disease areas) could all alter the applicability and suitability of the framework. It may, for example, in some circumstances be reasonably to run phases in parallel, but in others this might not be appropriate. In short, consideration should be given to how the framework can be tailored for different contexts (Miller, 1986).

Finally, it should be recognised that this framework is one small part of what will be a whole system transformation. It does not make sense to mandate the use of this framework, for example, until appropriate validation datasets are securely and widely available, and complex issues such as who is responsible for or the 'owner' of such datasets have been resolved. The characteristics of these datasets, and the design of the associated access models, will need to be planned carefully. Ideally, validation datasets will be accessible via securely designed trusted research environments, capable of handling high computational demand, which enable the data to remain in situ. Similarly, this framework, and other published guidelines, will not be sufficient to ensure public trust in the use of AI in healthcare settings. More will need to be done to ensure the process of AI design, development and deployment is provably trustworthy. This may entail, for example, greater open sharing of code, documentation and testing to reduce the extent to which the current AI development pipeline is inscrutable (Tsamados et al., 2020).

## Conclusion

The recent increase in interest from governments, and corresponding increase in public sector funding for the development of AI, is often driven by hype and can give the impression that using AI for healthcare is new. This impression is misleading and can be dangerous. It can delay policymakers, legislators and regulators from developing essential guardrails that can protect the public from harm. This synthesising exercise has shown that such delay, in the case of AI for health, is not justifiable on the basis of insufficient understanding of the technology. There is an extensive extant literature on the potential benefits and risks of AI for health, and considerable attention has been paid for decades to developing methods and guidelines for governing its use. The time has come to turn this knowledge into action.

## Acknowledgements

We would like to thank Emma Laycock, Emma Pencheon, Johan Ordish, Russell Pearson, Omar Moreea from NHSX, Medicines and Healthcare products Regulatory Agency (MHRA), and National Institute for Clinical Excellence (NICE), respectively, for their support throughout this project.